\documentclass[letterpaper, 10 pt, conference]{ieeeconf}  
\pdfoutput=1
\IEEEoverridecommandlockouts                              
\usepackage{verbatim}
\usepackage{graphics} 
\usepackage{amsmath,amssymb,stmaryrd,mathtools, amsfonts}
\usepackage{multirow}
\usepackage{hhline}
\usepackage{algorithm}
\usepackage[noend]{algpseudocode}
\usepackage{subfigure}
\usepackage{dsfont}
\usepackage{algorithmicx}
\usepackage{multirow}
\usepackage[utf8]{inputenc}
\usepackage{import}
\usepackage{graphicx, graphics}
\usepackage{array}
\usepackage{color,xcolor}
\usepackage{soul}

\newcommand{\AMax}[1]{\underset{#1}{\text{argmax}}\;}

\usepackage{siunitx} 

\title{\LARGE \bf
ParkPredict: Motion and Intent Prediction of Vehicles in Parking Lots
}

\author{Xu Shen$^{1,\star}$, Ivo Batkovic$^{2,3,\star}$, Vijay Govindarajan$^{1,\star}$, Paolo Falcone$^{2}$, Trevor Darrell$^{1}$, and Francesco Borrelli$^{1}$%
\thanks{This work is partly supported by Berkeley DeepDrive (BDD) and the Wallenberg Artificial Intelligence,
Autonomous Systems and Software Program (WASP) funded by
Knut and Alice Wallenberg Foundation.}%
\thanks{$^\star{}$ Indicates equal contribution.}
\thanks{$^{1}$ University of California, Berkeley, CA, USA (\{xu\_shen, govvijay, fborrelli, trevor\}@berkeley.edu).}%
\thanks{$^{2}$ Chalmers University of Technology, Gothenburg, Sweden (\{ivo.batkovic, falcone\}@chalmers.se).}%
\thanks{$^{3}$ Zenuity AB, Gothenburg, Sweden (ivo.batkovic@zenuity.com).}
}

\begin{document}
\maketitle
\thispagestyle{empty}
\pagestyle{empty}


\begin{abstract}
We investigate the problem of predicting driver behavior in parking lots, an environment which is less structured than typical road networks and features complex, interactive maneuvers in a compact space. Using the CARLA simulator, we develop a parking lot environment and collect a dataset of human  parking maneuvers. We then study the impact of model complexity and feature information by comparing a multi-modal Long Short-Term Memory (LSTM) prediction model and a Convolution Neural Network LSTM (CNN-LSTM) to a physics-based Extended Kalman Filter (EKF) baseline. Our results show that 1) intent can be estimated well (roughly 85\% top-1 accuracy and nearly 100\% top-3 accuracy with the LSTM and CNN-LSTM model); 2) knowledge of the human  driver’s  intended  parking  spot has a major impact on predicting parking trajectory; and 3) the semantic representation of the environment improves long term predictions.
\end{abstract}

\section{Introduction}
\label{sec:introduction}


While autonomous driving technologies have advanced by leaps and bounds, autonomous vehicles~(AVs) still face great challenges.  Robust perception~\cite{cv_for_avs_2017}, prediction~\cite{Lefevre2014}, and interaction in real traffic scenarios with other participants~\cite{Batkovic2019, driggs2017integrating}, especially human-driven vehicles, are difficult.
Depending on the estimated behavior of the others, an AV's behaviour may be too conservative, aggressive, or in the worst case, unsafe.

This problem is especially difficult in compact and unstructured domains like parking lots, which feature numerous interactions with human agents in close proximity~\cite{wired_parking} and may lead to congestion with suboptimal policies~\cite{Shen2019}. The potential to equip sensing and communicating infrastructure in these environments may help enable algorithmic solutions for connected AVs~\cite{Guanetti2018}.

Extensive research has been conducted to investigate the problem of vehicle motion prediction, and algorithms are developed upon various level of accessible information.

Physics-based methods use the pose history to provide intuitive short-term predictions~\cite{Houenou2013}, where Kalman Filters~\cite{schubert2008} are used to propagate the vehicle dynamics forward in time and predict a trajectory or reachable set~\cite{Althoff2009}.
On the other side, data-driven methods like Recurrent Neural Networks (RNNs), especially Long Short-Term Memory (LSTM) networks, can learn a motion model without explicit knowledge of the vehicle parameters or input profile. The encoder of the LSTM processes a pose history and produces a summary of this sequence, which is then passed to a decoder for prediction~\cite{Park2018,Kim2018}. 
\begin{figure}[t]
    \vspace{0.5em}
	\centering
	\subfigure[\small Bird's-eye view.]{
        \includegraphics[width=0.41\columnwidth]{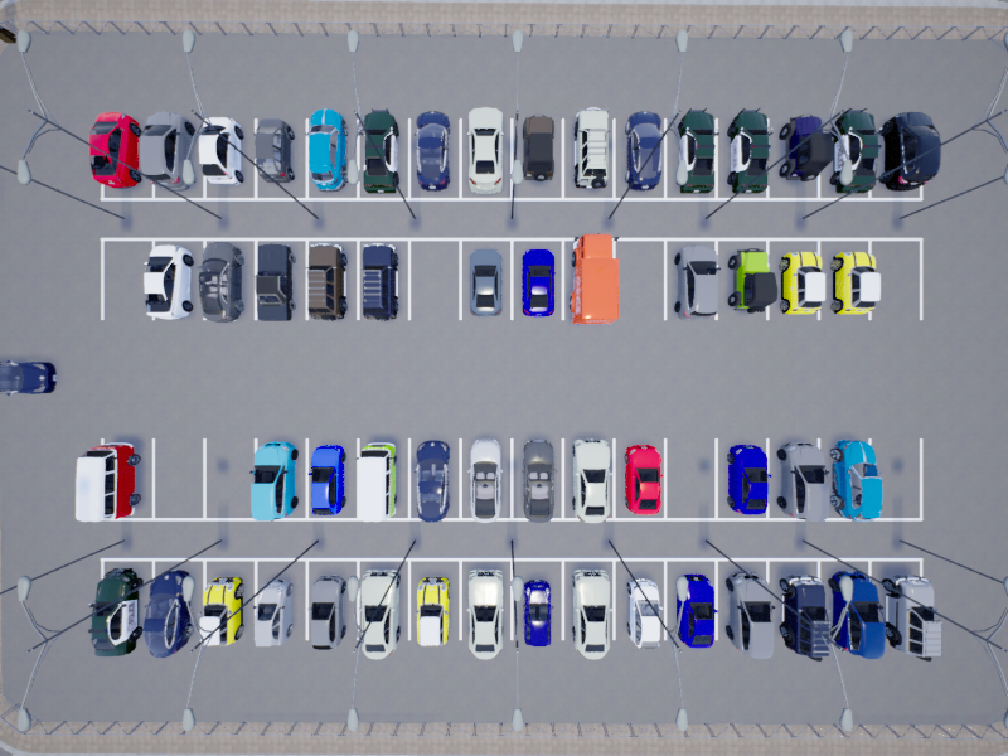}
        \label{fig:parking_lot}
    }
    \subfigure[\small Driver view.]{
        \includegraphics[width=0.5\columnwidth]{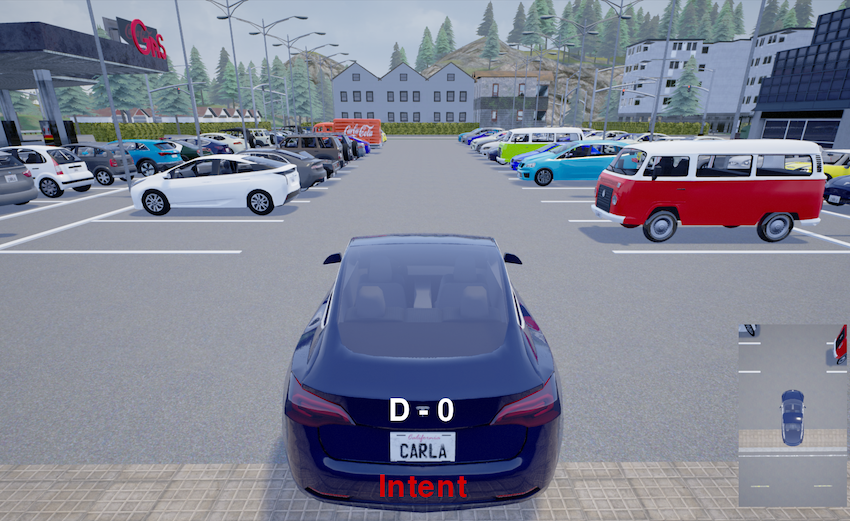}
        \label{fig:driver_view}
    }
	\caption{\small Parking lot scenario.}\vspace{-1em}
\end{figure}

Since the vehicle motion is directly influenced by the intent, such as turning, lane keeping, and lane changing, embedding the intent information with the motion information can improve prediction performance~\cite{prob_prediction_2018}. The LSTM encoder learns a representation of the motion data to predict
intent and generate interpretable multi-modal predictions~\cite{Deo2018}. 
%
%
For highway or urban driving, intent is usually classified based on lane graphs~\cite{Streubel2014, Schulz2018, Gindele2015}.

Not requiring explicit processing of trajectory or environment structure, Convolutional Neural Networks (CNNs) have been used in conjunction with LSTMs to synthesize information and make predictions from image sequences~\cite{donahue_lrcn_2015}.  In recent work, these architectures have been applied to bird's-eye view (BEV) semantic images~\cite{Casas2018}, including vehicle geometry, dynamics~\cite{cui2019deep}, and collision avoidance constraints for multimodal motion prediction~\cite{Cui2019}.

Most existing studies discussed above have focused on structured environments, e.g., there exists a well defined road network~\cite{vallon2017machine} consisting of intersections, lanes, and traffic lights. However, for environments such as parking lots, the following  challenges arise:
\begin{enumerate}
    \item only limited public datasets are available of human-driven vehicles inside parking lots~\cite{govea_ghmm};
    \item the parking maneuver is typically more complex~\cite{Zhang2019} and challenging than highway driving; 
    \item the compact space and proximity among surrounding objects increases the risk of collision and congestion.
\end{enumerate}

In this work, we focus on the problem of predicting a human driver's intended parking spot and future trajectory, given a set of features and levels of model abstraction. This paper offers the following contributions:
\begin{enumerate}
    \item we develop a parking lot simulation environment using the CARLA simulator and a racing wheel interface;
    \item we collect an annotated dataset of human parking behaviors including both forward and reverse parking maneuvers, as well as various parking spot selections;
    \item we propose a hierarchical LSTM model structure  which can provide both intent and multi-modal trajectory predictions;
    \item we propose a nested CNN-LSTM model structure with a visual encoder applied to semantic BEV images capturing parking lot geometry;
    \item we compare a physics-based Kalman Filter baseline against the higher complexity LSTM and CNN-LSTM models to investigate the impact of model complexity, feature complexity, and amount of accessible information on prediction performance.
\end{enumerate}

This paper is organized as follows.  Section \ref{sec:dataset} discusses the experimental design and dataset generation. Section \ref{sec:methods} elaborates on the algorithms designed for intent classification and motion prediction. Section \ref{sec:results} discusses the results of the prediction algorithms.  Finally, Section \ref{sec:discussion} summarizes our key findings and ideas for future work.

\section{Experimental Design and Dataset}
\label{sec:dataset}
This section provides an overview of the parking lot simulation environment and experimental setup. We generated a dataset from human driving demonstrations consisting of trajectories, final parked spot location, and signaled intent. This dataset was then used for intent and motion prediction.

\subsection{Parking Lot Scenario}\label{sec:methods_scenario}

In order to collect parking demonstrations in a controlled fashion, we used the CARLA simulator and CARLA ROS bridge \cite{Dosovitskiy17} with a custom parking lot map modified from Town04.  The parking lot consists of 4 rows with 16 spots each. In each trial, static vehicles were spawned into parking spots such that only 8 free spot options, located in the middle two rows, were available.  The specific locations of these free spots were varied across trials to gather a diverse range of parking demonstrations from the subject.  A free spot configuration example can be seen in Fig.~\ref{fig:parking_lot}.

Given this setting, the subject was instructed to park into a free spot of their choosing, following a specified forward or reverse parking maneuver.  When the subject selected the parking spot, he or she was instructed to press a button to signal a determined intent.  The subject used a Logitech G27 racing wheel to control brake, throttle, and steering of the ego vehicle.  In this experiment, only the ego vehicle was moving; all other vehicles in the scene remained static and parked.  The driver view is visualized in in Fig.~\ref{fig:driver_view}.

A total of $10$ subjects performed 30 forward parking and 30 reverse parking demonstrations, resulting in $600$ total demonstrations.  In each demonstration, the kinematic motion state history of the ego vehicle was recorded, as well as intent signals to know when a parking spot had been selected.  In addition, the locations of each parking spot was collected.  All demonstrations containing collisions or without intent signaling were filtered out. Furthermore, the configuration of the parking lot was recorded together with the bounding boxes of the ego vehicle and all other parked vehicles.
\subsection{Dataset Generation}\label{sec:methods_dataset}
\begin{figure}
    \centering
    \includegraphics[width=\linewidth]{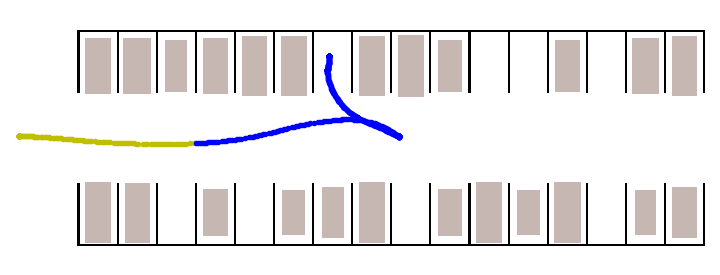}
    \caption{\small Sample demonstration focused on the middle two rows.  The shaded bounding boxes represent parked vehicles.  
    At first, the human driver has an undetermined intent (yellow section).  Then the driver decides the spot intent and trajectory is shown in blue.}
    \label{fig:demo}\vspace{-1em}
\end{figure}

We first introduce some notation used for the data included in each demonstration.  
\begin{itemize}
    \item The vehicle pose at time $t$ is denoted by
    $\vec{z}(t)= \begin{bmatrix} x(t) & y(t) & \theta(t) \end{bmatrix}^\intercal.$ We assume the vehicle lies on the $xy$\nobreakdash-plane and ignore variation in altitude, pitch, and roll.
    \item We denote the parking spot occupancy at time $t$ as
    $$\mathcal{O}(t) = \begin{bmatrix} x_{s,1}(t) & y_{s,1}(t) & f_{s,1}(t) \\ \vdots & \vdots & \vdots \\ x_{s,G}(t) & y_{s,G}(t) & f_{s,G}(t) \end{bmatrix},$$ which is a matrix of size $G$ by $3$, where $G$ is the number of parking spots.  In each row $j$, the first two entries, $x_{s,j}(t), y_{s,j}(t)$ are the spot location and the last entry, $f_{s,j}(t)$, is a binary variable set to 1 if the spot is free.
    
    \item The ground truth distribution of the driver intent is denoted by $g(t)$, which is a one-hot vector with size $G+1$. The $(G+1)$-th element corresponds to undetermined intent. After the intent is signaled, $g(t)$ corresponds to the spot where the subject finally parks in.

    
    \item The parking bird's-eye view $I(t)$ is a semantic image of shape $H$ by $W$ by $3$ representing the parking environment in Fig. \ref{fig:parking_lot}. The three channels correspond to the parking spot markings, static vehicle bounding boxes, and ego vehicle bounding box respectively. 
\end{itemize}
For each demonstration, as shown in Fig.~\ref{fig:demo}, the time intervals before the subject started driving and after the vehicle was parked were removed. 
Given a timestep $\Delta{}t=0.1$s, the remaining portion of demonstrations were processed into short snippets with a history horizon $N_\mathrm{hist}=5$, and a prediction horizon $N_\mathrm{pred}=20$\footnote{More details, as well as the processed dataset, can be found at https://bit.ly/parkpredict}. 
Each snippet was further processed to generate the following features:
\begin{enumerate}
    \item motion history of $\vec{z}(t)$: $\mathcal{Z}_\mathrm{hist}(t)\in\mathbb{R}^{N_\mathrm{hist}\times 3}$;
    \item parking spot and occupancy $\mathcal{O}(t)\in\mathbb{R}^{G\times{}3}$;
    \item image history of $I(t)$: $\mathcal{I}_\mathrm{hist}(t)\in\mathbb{R}^{N_\mathrm{hist}\times{}H\times{}W\times{}3}$; 
\end{enumerate}
and labels:
\begin{enumerate}
    \item future motion of $\vec{z}(t)$: $\mathcal{Z}_\mathrm{future}(t)\in\mathbb{R}^{N_\mathrm{pred}\times 3}$
    \item ground truth driver intent: $g(t)$.
\end{enumerate}
Note that all history features are sampled backward in time from $t$ with horizon $N_\mathrm{pred}$ and step size $\Delta{t}$.  Similarly, the prediction label is sampled forward in time from $t$ with horizon $N_\mathrm{future}$ and step size $\Delta{t}$.

This processing resulted in a dataset $\mathcal{D} = \{(\mathcal{Z}_\mathrm{hist}^{(i)}, \mathcal{O}^{(i)},\mathcal{I}_\mathrm{hist}^{(i)}), (\mathcal{Z}_\mathrm{future}^{(i)}, g^{(i)})  \}_{i=1}^{M}$.  Here, the superscript $(i)$ corresponds to the $i$-th dataset instance, and the total number of instances is $M=20850$.  The first tuple corresponds to a feature and the second tuple corresponds to a label, where we drop the time dependence of each entry going forward.

\section{Methodology}
\label{sec:methods}
Given the dataset $\mathcal{D}$, we break the prediction problem into intent and trajectory estimation subproblems.
Using input history $\mathcal{X}_\mathrm{hist}^{(i)}$ and occupancy $\mathcal{O}^{(i)}$, we generate the distribution of predicted driver intent $\hat{g}^{(i)}$ and the predicted future $N_\mathrm{pred}$-step trajectory $\hat{\mathcal{Z}}_\mathrm{future}^{(i)}$.  We evaluate three models that vary in both model and feature complexity: an EKF baseline, an LSTM network, and a CNN-LSTM network.  The following sections describe the model structures and input history features $\mathcal{X}_\mathrm{hist}^{(i)}$ provided to each model.

\subsection{Extended Kalman Filter (EKF) with Constant Velocity}\label{sec:methods_kf}

As a baseline, we use an EKF with a constant velocity assumption for which $\mathcal{X}_\mathrm{hist}^{(i)} = (\mathcal{Z}_\mathrm{hist}^{(i)})$.  The following state dynamics and measurement model are used.

\subsubsection{State Dynamics}
\begin{equation*}
    \begin{bmatrix}
    x_{k+1} \\
    y_{k+1} \\
    \theta_{k+1} \\
    v_{k+1} \\
    \omega_{k+1} \\
    \end{bmatrix}
    =
    \begin{bmatrix}
    x_k + v_k \cos( \theta_k ) \Delta t \\
    y_k + v_k \sin( \theta_k ) \Delta t \\
    \theta_k + \omega_k \Delta t \\
    v_k \\
    \omega_k \\
    \end{bmatrix}
    +
    q_k,
    ~
    q_k \sim \mathcal{N}(0, Q)
\end{equation*}
\subsubsection{Measurement Model}
\begin{equation*}
    \hat{\vec{z}}_k
    =
    \begin{bmatrix}
    1 & 0 & 0 & 0 & 0 \\
    0 & 1 & 0 & 0 & 0 \\
    0 & 0 & 1 & 0 & 0 \\
    \end{bmatrix}
    \begin{bmatrix}
    x_{k} \\
    y_{k} \\
    \theta_{k} \\
    v_{k} \\
    \omega_{k} \\
    \end{bmatrix}
    +
    r_k,
    ~
    r_k \sim \mathcal{N}(0, R)
\end{equation*}

The disturbance covariance, $Q$, was estimated using the pose and velocity information provided in each demonstration.  The noise covariance was chosen as $R=diag(1\text{e-}3, 1\text{e-}3, 1\text{e-}3)$, to reflect the ground truth pose measurement.  
We first run the time and measurement updates for the pose history $\mathcal{Z}_\mathrm{hist}$ and extrapolate $\hat{\mathcal{Z}}^{(i)}_{future}$ by the time update. 
Then, $\hat{g}^{(i)}$ is estimated by assigning probability to free spots (i.e. using $\mathcal{O}^{(i)}$)  based on the inverse of the Euclidean distance to the final predicted position.
%
If a given spot exceeds a distance threshold (20 m), then the corresponding probability is added to the undetermined category, the $(G+1)$-th element of $\hat{g}^{(i)}$.

\subsection{Long Short-Term Memory Network (LSTM)}\label{sec:methods_lstm}

\begin{figure*}[t]
\vspace{0.5em}
\centering
\includegraphics[width=\linewidth]{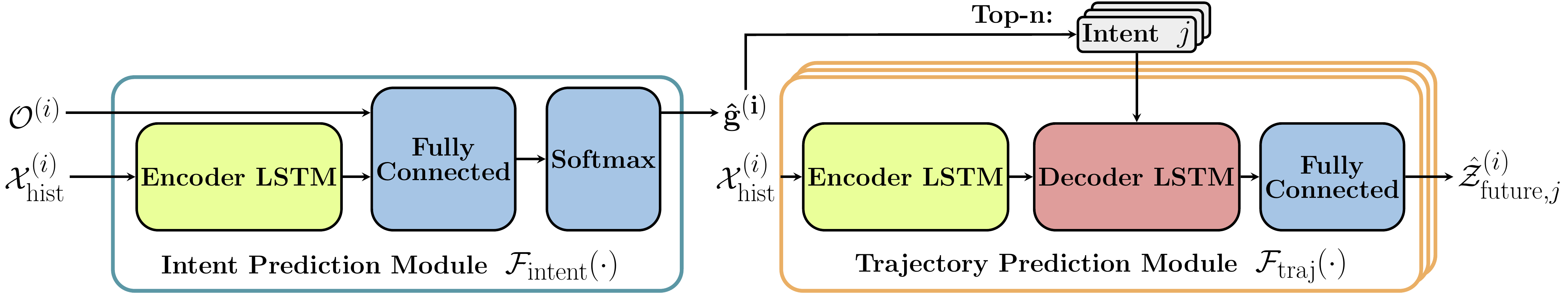}
\caption{\small Multi-modal LSTM prediction model.}
\label{fig:lstm_overall}
\vspace{-0.5em}
\end{figure*}

Our proposed LSTM model is shown in Fig.~\ref{fig:lstm_overall}. 
As with the EKF, $\mathcal{X}_\mathrm{hist}^{(i)} = (\mathcal{Z}_\mathrm{hist}^{(i)})$, but intent and trajectory estimation are addressed in the reversed order.  In particular, unlike the EKF, the model predicts multimodal, intent-conditioned trajectories with the following hierarchical structure.

\subsubsection{Intent Prediction Module}
The intent prediction module takes $\mathcal{X}_\mathrm{hist}^{(i)}$ and $\mathcal{O}^{(i)}$ as inputs and estimates the intent probability distribution:
\begin{equation*}
    \hat{g}^{(i)} = \mathcal{F}_{\mathrm{intent}}(\mathcal{X}_\mathrm{hist}^{(i)}, \mathcal{O}^{(i)})
\end{equation*}
%


The pose history and parking spot occupancy are first passed into the encoder LSTM stack and then processed by a fully connected layer.  A softmax output layer produces the predicted intent distribution, $\hat{g}^{(i)}$.

The objective function we minimize for intent prediction has three components, where $j \in \{1,...,G+1\}$ denotes the intent index.
\begin{itemize}
    \item Cross entropy between prediction $\hat{g}^{(i)}$ and ground truth $g^{(i)}$ to drive the predicted distribution closer to the ground truth label:
    \begin{equation*}
        J_1^{\mathrm{intent}} = 
        - \sum_{j=1}^{G+1} g^{(i)}_j \log(\hat{g}^{(i)}_j).
    \end{equation*}
    \item Negative entropy of the predicted distribution $\hat{g}^{(i)}$ to account for the stochasticity of the driver:
    \begin{equation*}
        J_2^{\mathrm{intent}} = - \mathcal{H}(\hat{g}^{(i)}) = \sum_{j=1}^{G+1} \hat{g}^{(i)}_j \log(\hat{g}^{(i)}_j).
    \end{equation*}
    \item Penalty of predicting already occupied parking spots:
    \begin{equation*}
        J_3^{\mathrm{intent}} = \sum_{j=1}^{G} \max\{(\hat{g}^{(i)}_j - f_{s,j}^{(i)}), 0 \}.
    \end{equation*}
\end{itemize}
Therefore, the final objective function is
\begin{equation*}
    J^{\mathrm{intent}} = J_1^{\mathrm{intent}} + 
    J_2^{\mathrm{intent}} + J_3^{\mathrm{intent}}.
\end{equation*}

\subsubsection{Trajectory Prediction Module}
The trajectory prediction module takes $\mathcal{X}_\mathrm{hist}^{(i)}$ and the intent index $j \in \{1,...,G+1\}$ as input and estimates the future trajectory for $N_{pred}$ timesteps:
\begin{equation*}
    \hat{\mathcal{Z}}_{\mathrm{future},j}^{(i)} = \mathcal{F}_\mathrm{traj}(\mathcal{X}_\mathrm{hist}^{(i)}, j).
\end{equation*}
This module works sequentially with the intent prediction module to generate multimodal predictions. The encoder first processes $\mathcal{X}_\mathrm{hist}^{(i)}$. The encoder's final hidden state and cell state are used to initialize the decoder. Then, for each intent index $j$, the decoder and the subsequent fully connected layer return a predicted future trajectory $\hat{\mathcal{Z}}_{\mathrm{future},j}^{(i)}$, which is associated with probability $\hat{g}^{(i)}_j$.


However, during training, we decouple this module with intent prediction and only use ground truth label $g^{(i)}$: 
\begin{equation*}
    \hat{\mathcal{Z}}_{\mathrm{future},gt}^{(i)} = \mathcal{F}_\mathrm{traj}(\mathcal{X}_\mathrm{hist}^{(i)}, \AMax{j} g^{(i)}_j).
\end{equation*}
 The objective function $J^{\mathrm{traj}}$ is defined using mean squared error (MSE) on position.  Concretely, let $\vec{z}_k^{\>(i)}$ and $\hat{\vec{z}}_k^{\>(i)}$ be the $k$-th row of $\mathcal{Z}_\mathrm{future}^{(i)}$ and $\hat{\mathcal{Z}}_\mathrm{future}^{(i)}$ respectively, then:
\begin{equation*}
    \small
    J^{\mathrm{traj}} = \frac{1}{N_\mathrm{pred}}\hspace{-0.4em} \sum_{k=1}^{N_\mathrm{pred}}\hspace{-0.4em}  
    \sqrt{
        (\vec{z}_k^{\>(i)}\hspace{-0.2em}-\hspace{-0.1em}\hat{\vec{z}}_k^{\>(i)})^\top 
        \mathrm{diag}(1,1,0)
        (\vec{z}_k^{\>(i)}\hspace{-0.2em}-\hspace{-0.1em}\hat{\vec{z}}_k^{\>(i)})
    }   
\end{equation*}

\subsection{Convolutional Neural Network LSTM (CNN-LSTM)}\label{sec:methods_cnn}

\begin{figure}
	\centering
	\subfigure{
        \includegraphics[width=0.32\columnwidth]{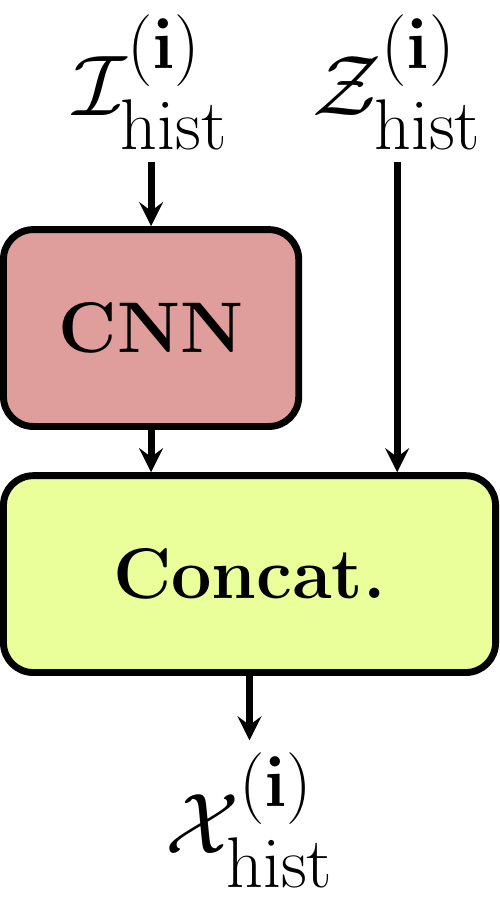}
    }
    \subfigure{
        \includegraphics[width=0.58\columnwidth]{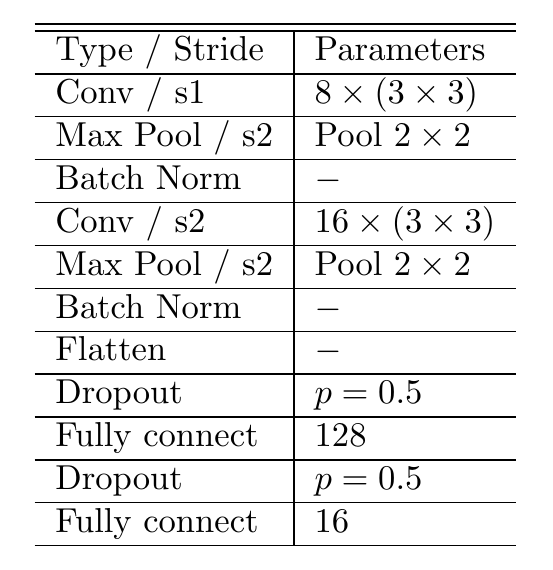}
    }
	\caption{\small CNN architecture. The table on the right describes the CNN preprocessing block in the left figure.}\label{fig:cnn_preprocessing}\vspace{-1em}
\end{figure}
The CNN-LSTM model, like the LSTM model, uses the same structure in Fig.~\ref{fig:lstm_overall}.  However, as depicted in Fig.~\ref{fig:cnn_preprocessing}, a single CNN preprocesses each image in $\mathcal{I}^{(i)}_\mathrm{hist}$.  The generated image features are then concatenated with the motion features. Hence, for the CNN-LSTM model, $\mathcal{X}_\mathrm{hist}^{(i)} = (\mathcal{F}_\mathrm{CNN}(\mathcal{I}_\mathrm{hist}^{(i)}), \mathcal{Z}_\mathrm{hist}^{(i)})$, where $\mathcal{F}_\mathrm{CNN}(\cdot)$ represents the visual encoding operation performed by the CNN.  This is inspired from the approaches used in~\cite{Cui2019, bdd_end_to_end_2017}, which fuse motion and visual features with a LSTM temporal encoder.

\subsection{Prediction Evaluation}
To compare the performance of each prediction algorithm, we use 5-fold cross validation, where the LSTM and CNN-LSTM are trained for 200 epochs with batch size 32.
The variable $\tilde{M}$ here corresponds to the cardinality of the corresponding hold-out set being evaluated.
\begin{enumerate}
    \item Top-$n$ Accuracy:  Let the set $\mathcal{G}_n(\hat{g}^{(i)})$ include the $n$ most likely intent categories in the predicted intent distribution, $\hat{g}^{(i)}$.  Then, the top-$n$ accuracy is:
    \begin{equation*}
    \begin{array}{ll}
    \mathcal{A}_n = \frac{1}{\tilde{M}}\sum_{i=1}^{\tilde{M}} \sum_{j=1}^{G+1} ~ g_j^{(i)} ~ \mathbb{I}\left(  g_j^{(i)} \in \mathcal{G}_n(\hat{g}^{(i)}) \right)
    \end{array}
    \end{equation*}
    where $\mathbb{I}(\cdot)$ is the indicator function.
    \item Mean Distance Error: Given a predicted  $\hat{\mathcal{Z}}_{future}$ and actual trajectory $\mathcal{Z}_{future}$, we can look at the position error as a function of the prediction timestep.  The mean distance error at timestep $k$, $d_k$, is: 
    \begin{equation*}
    \begin{array}{ll}
        d_k = \\  \frac{1}{\tilde{M}} \sum_{i=1}^{\tilde{M}} 
        \sqrt{
        (\vec{z}_k^{\>(i)} - \hat{\vec{z}}_k^{\>(i)})^\top 
        ~
        diag(1,1,0)
        ~
        (\vec{z}_k^{\>(i)} - \hat{\vec{z}}_k^{\>(i)})
        }
    \end{array}
    \end{equation*}
\end{enumerate}

\section{Results}


\label{sec:results}
In this section, we compare the intent prediction capability and the trajectory prediction error of each algorithm using the selected evaluation metrics.  We investigate the impact of information level and multimodality
on the prediction performance.  For brevity, CNN-LSTM is shortened to CNN in the subsequent figures.

\subsection{Intent Classification}


\begin{figure}[ht]
    \centering
    \includegraphics[width=\linewidth]{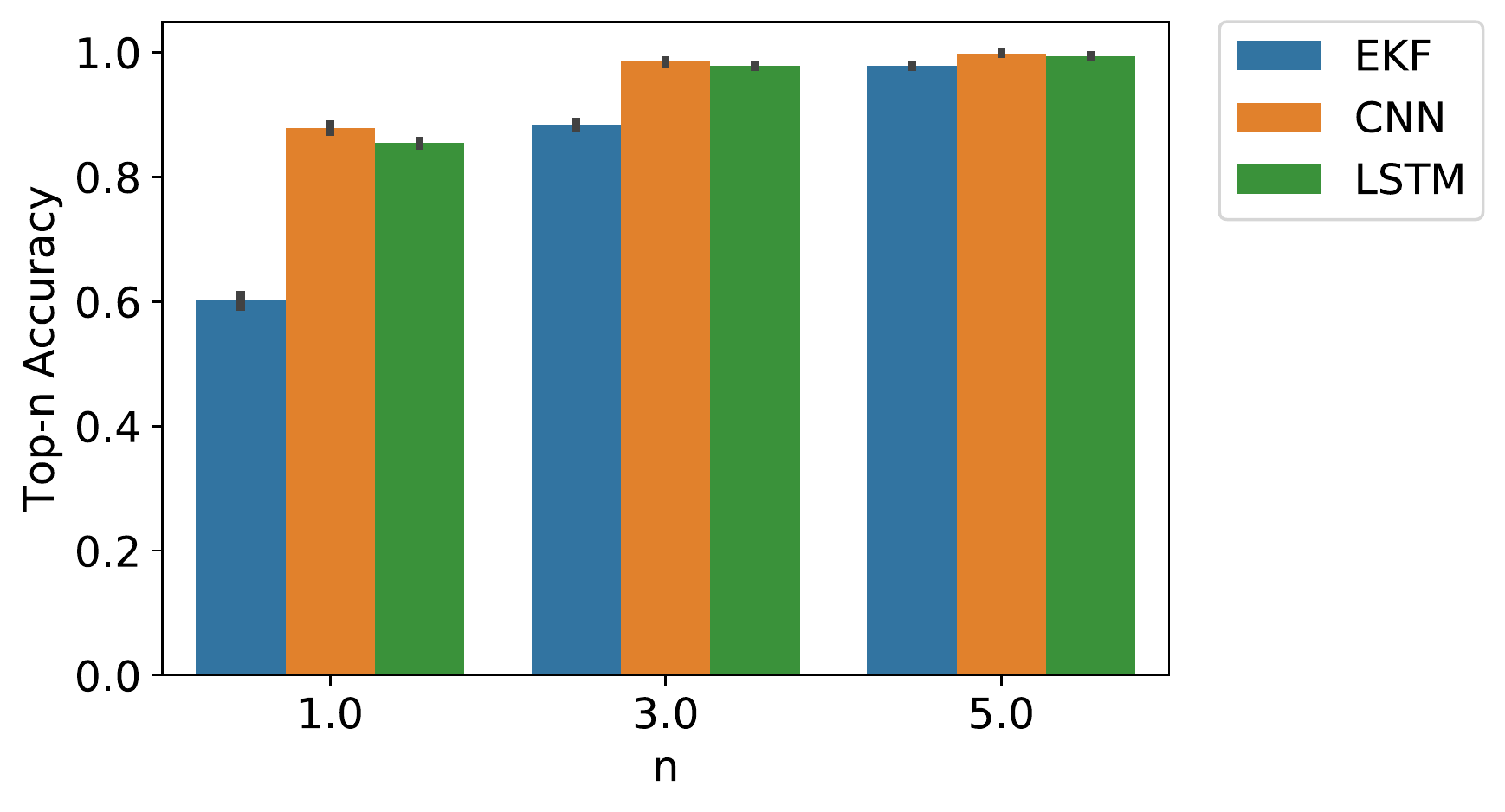}
    \caption{\small Top-n intent classification accuracy, $\mathcal{A}_n$.}
    \label{fig:intent_acc}
\end{figure}

Fig.~\ref{fig:intent_acc} shows the top-$n$ accuracy $\mathcal{A}_n$ across all models evaluated.  As expected, the LSTM and CNN-LSTM outperforms the EKF baseline at every $n$, achieving roughly $85$\% top-1 accuracy and nearly 100\% top-3 or top-5 accuracy.  This follows from the fact that the LSTM and CNN-LSTM are trained specifically on the intent prediction task, while a heuristic inverse distance approach is used after the trajectory prediction step of the EKF.  

\subsection{Trajectory Prediction}
%
\begin{figure}[hb]
    \centering
    \vspace{-1em}
    \includegraphics[width=\linewidth]{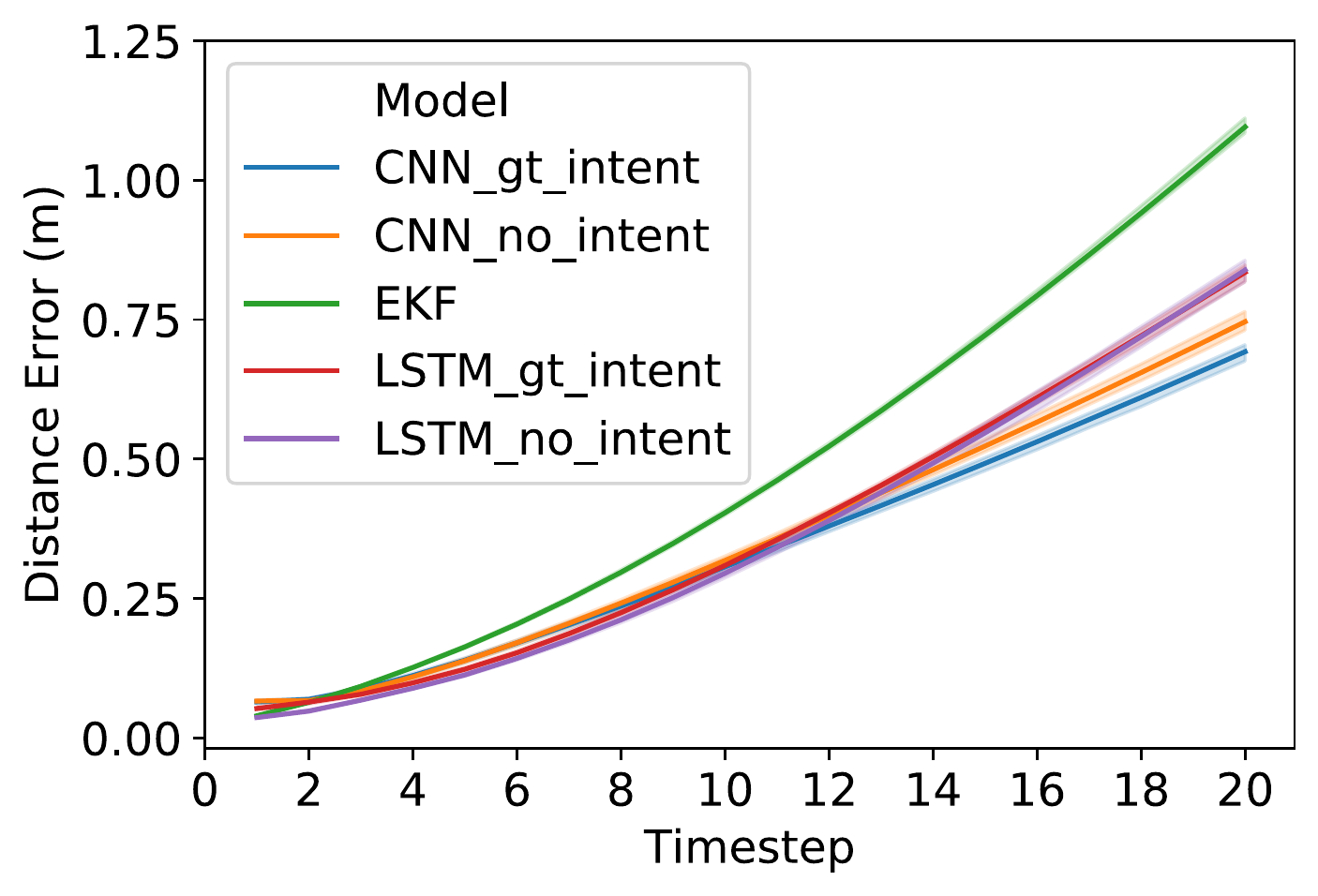}
    \caption{\small Mean distance error $d_k$ vs. timestep $k$ across varying models and levels of intent knowledge.}
    \label{fig:traj_error_models_intent}
\end{figure}

\begin{figure}[ht]
    \centering
    \includegraphics[width=\linewidth]{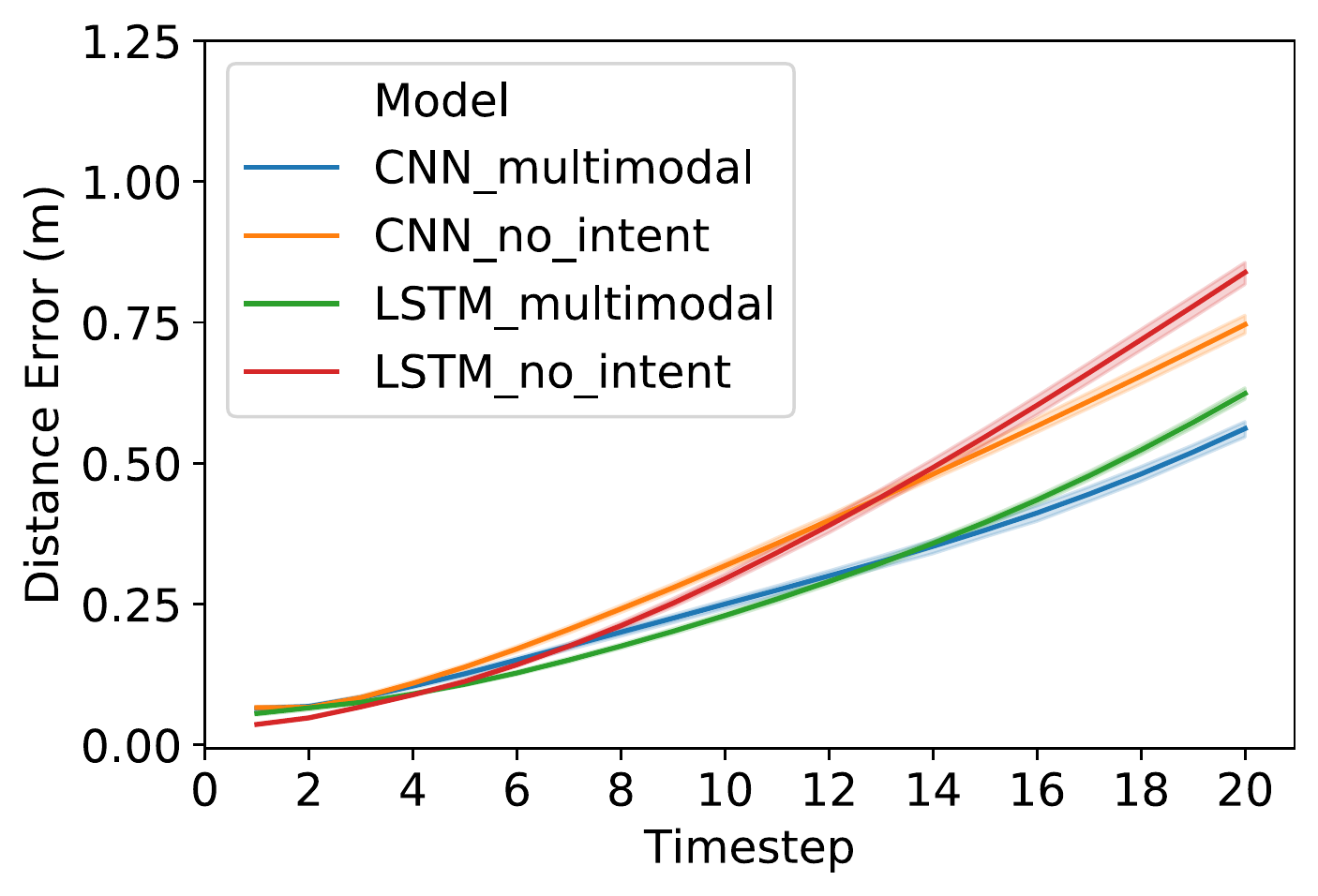}
    \caption{\small Mean distance error $d_k$ vs. timestep $k$ comparing intent-agnostic unimodal predictions to the best rollout among multimodal, intent-conditioned predictions.}
    \label{fig:traj_error_models_multimodal}
\end{figure}


In this section, we look at the following models and levels of information, where $\star \in \{\mathrm{LSTM}, \mathrm{CNN}\}$:
\begin{itemize}
    \item \textbf{$\star$\_no\_intent}: intent-agnostic model where the trajectory module $\mathcal{F}_{\mathrm{traj}}(\cdot)$ is re-trained and applied only zeroed intent input.
    \item \textbf{$\star$\_gt\_intent}: intent-conditioned model where $\mathcal{F}_{\mathrm{traj}}(\cdot)$ is decoupled from intent module $\mathcal{F}_{\mathrm{intent}}(\cdot)$, predicting $\hat{\mathcal{Z}}_{future,gt}^{(i)}$ only using ground truth intent $g^{(i)}$.
    \item \textbf{$\star$\_multimodal}: intent-conditioned, multimodal model where $\mathcal{F}_{\mathrm{traj}}(\cdot)$ is applied to the top-$n$ entries of $\hat{g}^{(i)}$ from $\mathcal{F}_{\mathrm{intent}}(\cdot)$.  
    Here we select $n = 3$.
\end{itemize}

Fig.~\ref{fig:traj_error_models_intent} captures the impact of intent knowledge and information level on the mean distance error, $d_k$, at each timestep $k$.  
For both the CNN-LSTM and LSTM models, we observe that they outperform the EKF for long-term predictions, as they learn a more nuanced motion model. For the LSTM model, the benefit of intent knowledge is  minimal, as the no intent model is on par with the ground truth model. However, in the case of the CNN-LSTM, the additional knowledge of intent aids the prediction, as seen after time step $12$. This suggests that the semantic BEV images can help the model to better understand the intent label for prediction.



Fig.~\ref{fig:traj_error_models_multimodal} shows the benefit of multimodal predictions on the mean distance error, $d_k$, at each timestep $k$.  Note that for the multimodal predictions, we take the top $n=3$ intent labels and generate 3 separate rollouts. The results reported here are computed by finding the rollout nearest the ground truth trajectory (i.e. using mean squared error on position, $J^{\mathrm{traj}}$) and then using this single rollout to evaluate $d_k$.
We note that every intent-conditioned model does on par or outperforms the model without intent knowledge, over the prediction horizon.  That shows that even if the intent is predicted and not known precisely, this additional information can still reduce trajectory prediction error compared to intent-agnostic predictions.  This is likely due, in part, to the multimodal nature of the trajectory prediction, which can capture more evolution possibilities and driver model stochasticity relative to a unimodal trajectory prediction.  We also observe that the CNN-LSTM models are well-suited for long-term predictions, for which the geometry of the parking lot is likely more informative.  




\subsection{Scenario Examples}

\begin{figure}[ht!]
    \centering
    \includegraphics[width=\linewidth]{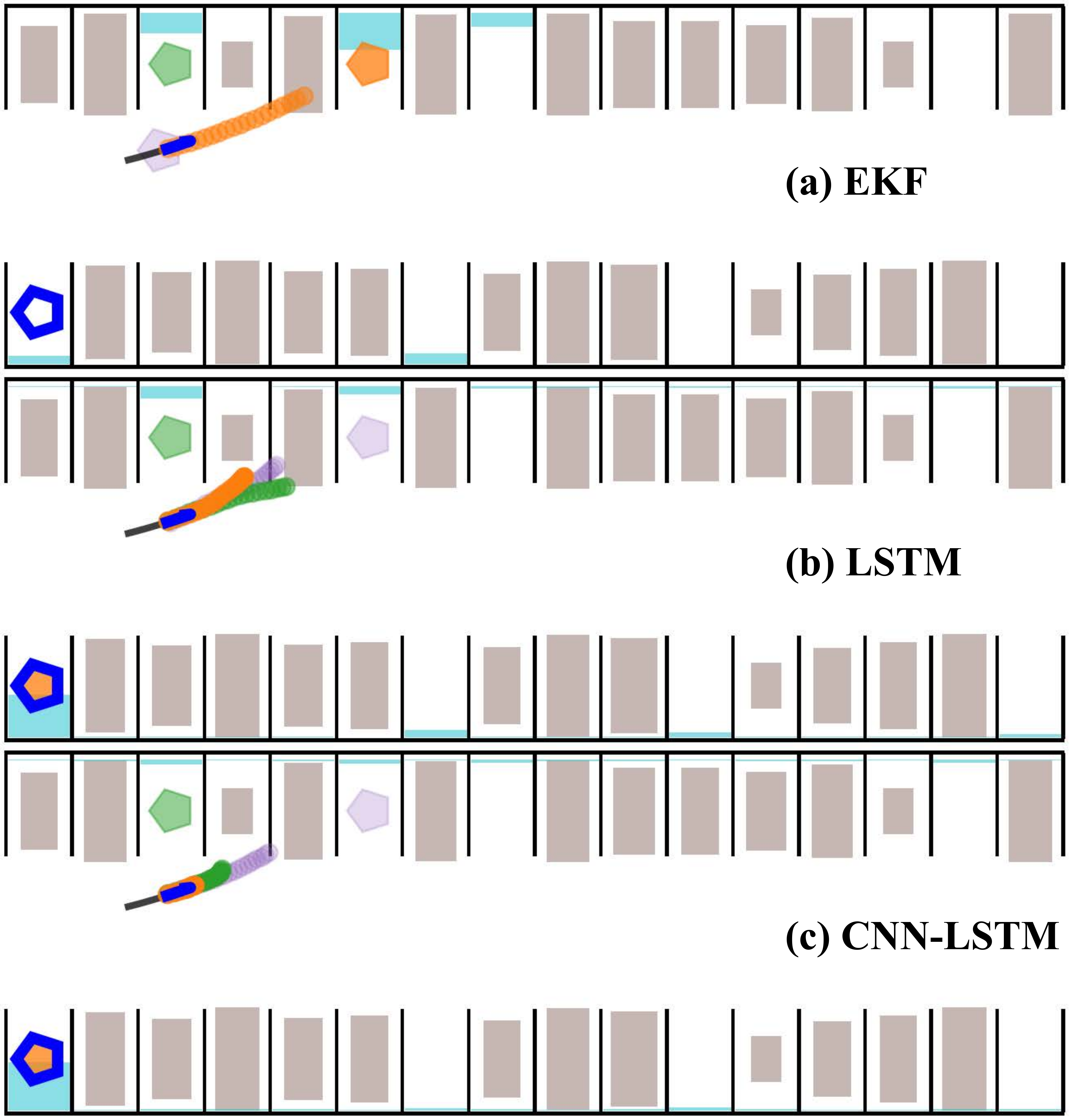}
    \caption{\small Prediction example across models. The black curve represents the pose history $\mathcal{Z}_{hist}^{(i)}$; the blue pentagon and curve stand for ground truth future intent $g^{(i)}$ and motion $\mathcal{Z}_{future}^{(i)}$ respectively; the orange, green, purple pentagons and curves  correspond to the top-3 intent and trajectory predictions. Their order and transparency levels reflect the probability $\hat{g}^{(i)}_j$, which is visualized by the cyan bars in spots. When a pentagon is marked on the trajectory, it corresponds to the undetermined intent.}
    \label{fig:scenario_example}\vspace{0em}
\end{figure}

We show how the prediction algorithms compare for a sample prediction instance in Fig.~\ref{fig:scenario_example}. The three sub-figures depict the 2-D layout of the parking lot, where the filled shaded boxes are occupied parking spots.
%
%


For intent prediction, the proximity-based heuristics of the EKF prioritizes the nearest spots in front of the vehicle, but misses the case that the vehicle may reverse into the ground truth spot backwards. Instead, the LSTM and CNN-LSTM capture the ground truth in top-3 candidates, meaning that they also learned from data that the driver might choose to back up in their maneuver.

For trajectory prediction, the EKF extrapolates the dynamics so it only offers single trajectory prediction with a significant delay. The LSTM and CNN-LSTM fit the ground truth better and offer multi-modal behaviors for other likely nearby spots. The top-3 candidates from the LSTM and CNN-LSTM are relatively more aware of the obstacles, as the LSTM better leverages the occupancy information and the CNN-LSTM learns a more detailed obstacle representation through semantic image inputs. Therefore, both models place more emphasis on the reverse maneuver.
\section{Discussion}
\label{sec:discussion}

This work investigated the problem of predicting a human driver's parking intent and maneuver with varying levels of feature and model complexity.  A custom CARLA simulator parking lot environment was constructed and used to generate a dataset of human parking maneuvers.  We compared an intent-conditioned LSTM and CNN-LSTM prediction model against an EKF baseline and noted the benefits of providing intent information for trajectory prediction, even if estimated. Additionally, by encoding obstacles and parking lot geometry, the semantic BEV images help improve prediction performance in the long term. The hierarchical framework is capable of offering multi-modal driver behavior prediction in a relatively unstructured environment like parking lots. 

For future work, we would like to investigate the cost function used for the CNN-LSTM model by incorporating a penalty according to kinematic constraints as in \cite{cui2019deep}. In addition, by expanding the dataset to collect a wider variation of real world behaviors, we hope to apply the multimodal predictions in a stochastic control framework for autonomous parking in multi-agent settings.





\bibliographystyle{IEEEtran}

\bibliography{ref.bib}

 

\end{document}